\pdfoutput=1
\documentclass[11pt]{article}

\usepackage{emnlp2021}

\usepackage{times}
\usepackage{latexsym}

\usepackage[T1]{fontenc}
\usepackage[utf8]{inputenc}

\usepackage{microtype}

\usepackage{multirow}
\usepackage{bm}
\usepackage{amsmath}
\usepackage{mathtools}
\usepackage{subfig}
\usepackage{graphicx}
\usepackage{algorithm}
\usepackage{algpseudocode}
\usepackage{hyperref}
\usepackage{tikz}
\usepackage{amssymb}
\usepackage{fancyvrb}
\usepackage{xcolor}

\newcommand{\argmax}{\mathop{\mbox{argmax}}}
\newcommand{\argmin}{\mathop{\mbox{argmin}}}

\def\va{\mathbf{a}}
\def\vb{\mathbf{b}}

\title{From Simultaneous to Streaming Machine Translation by Leveraging Streaming History}

\author{Javier Iranzo-S\'{a}nchez \and  Jorge Civera \and   Alfons Juan \\
Machine Learning and Language Processing Group\\
Valencian Research Institute for Artificial Intelligence\\
Universitat Polit\`{e}cnica de Val\`{e}ncia\\
Cam\'{\i} de Vera s/n, 46022 Val\`{e}ncia, Spain \\
  \texttt{ \{jairsan,jorcisai,ajuanci\}@vrain.upv.es  }\\}

\begin{document}
\maketitle
\begin{abstract}
Simultaneous Machine Translation is the task of incrementally
translating an input sentence before it is fully available. Currently,
simultaneous translation is carried out by translating each sentence
independently of the previously translated text. More generally,
Streaming MT can be understood as an extension of Simultaneous MT to
the incremental translation of a continuous input text stream. In this
work, a state-of-the-art simultaneous sentence-level MT system is
extended to the streaming setup by leveraging the streaming
history. Extensive empirical results are reported on IWSLT Translation
Tasks, showing that leveraging the streaming history leads to
significant quality gains. In particular, the proposed system proves
to compare favorably to the best performing systems.
\end{abstract}

\section{Introduction}

Simultaneous Machine Translation (MT) is the task of incrementally
translating an input sentence before it is fully available.  Indeed,
simultaneous MT can be naturally understood in the scenario of
translating a text stream as a result of an upstream Automatic Speech
Recognition (ASR) process. This setup defines a simultaneous Speech
Translation (ST) scenario that is gaining momentum due to the vast
number of industry applications that could be exploited based on this
technology, from person-to-person communication to subtitling of
audiovisual content, just to mention two main applications.

These real-world streaming applications motivate us to move from
simultaneous to streaming MT, understanding streaming MT as the task
of simultaneously translating a potentially unbounded and unsegmented
text stream. Streaming MT poses two main additional challenges over
simultaneous MT. First, the MT system must be able to leverage the
streaming history beyond the sentence level both at training and
inference time. Second, the system must work under latency constraints
over the entire stream.

With regard to exploiting streaming history, or more generally
sentence context, it is worth mentioning the significant amount of
previous work in offline MT at sentence level~\cite{Tiedemann2017,Agrawal2018},
document level~\cite{Scherrer2019,Ma2020,Zheng2020b,Li2020b,Maruf2021,zhang-etal-2021-beyond},
and in related areas such as language modelling~\cite{Dai2019} that
has proved to lead to quality gains. Also, as reported in~\cite{Li2020b},
more robust ST systems
can be trained by taking advantage of the context across sentence
boundaries using a data augmentation strategy similar to the prefix
training methods proposed in~\cite{Niehues2018,Ma2019}.
This
data augmentation strategy was suspected to
boost re-translation performance when compared to
conventional simultaneous MT systems~\cite{Arivazhagan2020b}.

Nonetheless, with the notable exception of~\cite{Schneider2020},
sentences in simultaneous MT are still translated independently from
each other ignoring the streaming history.~\cite{Schneider2020}
proposed an end-to-end streaming MT model with a Transformer
architecture based on an Adaptive Computation Time method with a
monotonic encoder-decoder attention.  This model successfully uses the
streaming history and a relative attention mechanism inspired by
Transformer-XL~\cite{Dai2019}. Indeed, this is an MT model that
sequentially translates the input stream without the need for a
segmentation model. However, it is hard to interpret the latency
of their streaming MT model because the authors observe that 
the current
sentence-level latency measures, Average Proportion
(AP)~\cite{ChoE16E}, Average Lagging (AL)~\cite{Ma2019} and
Differentiable Average Lagging (DAL)~\cite{Cherry2019} do not perform
well on a streaming setup. This fact is closely related to the second
challenge mentioned above, which is that the system must work under latency
constraints over the entire stream. Indeed, current sentence-level 
latency measures
do not allow us to appropriately gauge the latency of streaming MT
systems. To this purpose,~\cite{Iranzo2021stream} recently proposed a
stream-level adaptation of the sentence-level latency measures based
on the conventional re-segmentation approach applied to the ST output
in order to evaluate translation quality~\cite{MatusovLBN05}.

In this work, the simultaneous MT model based on a unidirectional
encoder-decoder and training along multiple wait-$k$ paths proposed
by~\cite{Elbayad2020} is evolved into a streaming-ready simultaneous
MT model. To achieve this, model training is performed following a
sentence-boundary sliding-window strategy over the parallel stream
that exploits the idea of prefix training, while inference is carried
out in a single forward pass on the source stream that is segmented by
a Direct Segmentation (DS) model~\cite{Iranzo2020b}. In addition, a
refinement of the unidirectional encoder-decoder that takes advantage
of longer context for encoding the initial positions of the streaming
MT process is proposed. This streaming MT system is thoroughly
assessed on IWSLT translation tasks to show how leveraging the
streaming history provides systematic and significant BLEU
improvements over the baseline, while reported stream-adapted latency
measures are fully consistent and interpretable.  Finally, our system
favourably compares in terms of translation quality and latency to the
latest state-of-the-art simultaneous MT systems~\cite{Ansari2020}.

This paper is organized as follows. Next section provides a formal
framework for streaming MT to accommodate streaming history in
simultaneous MT.  Section~\ref{sec:exp} presents the streaming
experimental setup whose results are reported and discussed in
Section~\ref{sec:eva}. Finally, conclusions and future work are drawn
in Section~\ref{sec:con}.

\section{Streaming MT}
\label{sec:streaming}

In streaming MT, the source stream $X$ to be translated into $Y$ comes
as an unsegmented and unbounded sequence of tokens. In this setup, the
decoding process usually takes the greedy decision of which token
appears next at the $i$-th position of the translation being generated
\begin{equation}\label{eqn:streaming_mt}
\hat Y_i = \argmax\limits_{y \in \mathcal{Y}} p \!\left( y \bigm| X_1^{G(i)}, Y_1^{i-1} \right)
\end{equation}
where $G(i)$ is a global delay function that tells us the last
position in the source stream that was available when the $i$-th target
token was output, and $\mathcal{Y}$ is the target vocabulary.
However, taking into account the entire source and target streams can
be prohibitive from a computational viewpoint, so the generation of
the next token can be conditioned to the last $H(i)$ tokens of the
stream as
\begin{equation}\label{eqn:streaming_mt_history}
\hat Y_i \!=\! \argmax\limits_{y \in \mathcal{Y}} p \! \left( y \bigm| X_{G(i)-H(i)+1}^{G(i)}, Y_{i-H(i)}^{i-1} \right).
\end{equation}
Nevertheless, for practical purposes, the concept of sentence
segmentation is usually introduced to explicitly indicate a monotonic
alignment between source and target sentences in streaming MT. Let us
consider for this purpose the random variables $\va$ and $\vb$ for the
source and target segmentation of the stream, respectively. Variables
$\va$ and $\vb$ can be understood as two vectors of equal length
denoting that the $n$-th source sentence starts at position $a_n$,
while the $n$-th target sentence does so at position $b_n$.

In the next sections, we reformulate simultaneous MT in terms of the more
general framework of streaming MT. This reformulation allows us to
consider opportunities for improvement of previous simultaneous MT
models.

% we can use the notion of sentence segmentation to limit the
% history. Thus, we have a history function $H(i)$, that limits the
% number of tokens that are considered at each decoding step.
% \begin{equation}
% \hat Y_i = \argmax\limits_{y \in \mathcal{Y}}
% p \left( y | X_{a_{H(i)}}^{G(i)}, Y_{b_{H(i)}}^{i-1} \right)
% \end{equation}

% Similarly, $b_n$ marks the start of the $n$-th target sentence.
% $|a|=|b|=N$, and $a_1=1$, $b_1=1$. During training, the segmentation
% is known.

\subsection{Simultaneous MT with streaming history}
\label{sec:history}

In the conventional simultaneous MT setup, the aforementioned
variables $\va$ and $\vb$ are uncovered at training and inference
time, while in streaming MT $\va$ and $\vb$ are considered hidden
variables at inference time that may be uncovered by a segmentation
model. In fact, in conventional simultaneous MT the history is limited
to the current sentence being translated,
% that is, $H(i)=i-a_n$ with $a_n \leq i < a_{n+1}$, 
while in streaming MT we could exploit the fact that the history could
potentially span over all the previous tokens before the current
sentence.

To this purpose, the global delay function $G(i)$ introduced above
would replace the sentence-level delay function $g(i)$ commonly used
in simultaneous MT. However, it should be noticed that we could
express $g(i)$ as $G(i)-a_n$ with $b_n \leq i < b_{n+1}$.  Delay
functions are defined as a result of the policy being applied.  This
policy decides what action to take at each timestep, whether to read a
token from the input or to write a target token.  Policies can be
either fixed~\cite{Ma2019,Dalvi2018} depending only on the current
timestep, or adaptive~\cite{Arivazhagan2019,MaPCPG20,Zheng2020}
being also conditioned on the available input source words. Among
those fixed policies, the sentence-level wait-k policy proposed
by~\cite{Ma2019} is widely used in simultaneous MT with the simple
local delay function
\begin{equation}
\label{eqn:gi}
g(i)= k + i - 1. 
\end{equation}

This policy initially reads $k$ source tokens without writing a target
token, and then outputs a target token every time a source token is
read. This is true in the case that the ratio between the source and
target sentence lengths is one. However, in the general case, a
catch-up factor $\gamma$ computed as the inverse of the source-target
length ratio defines how many target tokens are written for every read
token, that generalises Eq.~\ref{eqn:gi} as
% 
% Hay que usar con el asterisco
\DeclarePairedDelimiter\ceil{\lceil}{\rceil}
\DeclarePairedDelimiter\floor{\lfloor}{\rfloor}
\begin{equation}
g(i)= \left \lfloor k + \frac{i-1}{\gamma} \right \rfloor.
\end{equation}

The wait-$k$ policy can be reformulated in streaming MT 
so that the wait-$k$ behaviour is carried out for each sentence as
\begin{equation}
G(i)= \left \lfloor k + \frac{i-b_n}{\gamma} \right \rfloor + a_n-1
\end{equation}
where $b_n \leq i < b_{n+1}$. 

In streaming MT, we could take advantage of the streaming history by
learning the probability distribution stated in
Eq.~\ref{eqn:streaming_mt_history}, whenever streaming samples would
be available. However, training such a model with arbitrarily long
streaming samples poses a series of challenges that need to be
addressed.  Firstly, it would be necessary to carefully define $G(i)$
and $H(i)$ functions so that, at each timestep, the available source
and target streams are perfectly aligned.  Given that the
source-target length ratio may vary over the stream, if one uses a
wait-$k$ policy with a fixed $\gamma$, there is a significant chance
that source and target are misaligned at some points over the stream.
Secondly, every target token can potentially have a different $G(i)$
and $H(i)$, so the encoder-decoder representation and contribution to
the loss would need to be recomputed for each target token at a
significant computational expense.  Lastly, current MT architectures
and training procedures have evolved conditioned by the availability
of sentence-level parallel corpora for training, so they need to be
adapted to learn from parallel streams.

To tackle the aforementioned challenges in streaming MT, a compromise
practical solution is to uncover the source and target sentence
segmentations. At training time, parallel samples are extracted by a
sentence-boundary sliding window spanning over several sentences of
the stream that shifts to the right one sentence at a time. In other
words, each sentence pair is concatenated with its corresponding
streaming history that includes previous sentence pairs simulating
long-span prefix training. Doing so, we ensure that source and target
streams are properly aligned at all times, and training can be
efficiently carried out by considering a limited history.  The
inference process is performed in a purely streaming fashion in a
single forward pass as defined in Eq.~\ref{eqn:streaming_mt_history}
with $H(i)$ being consistently defined in line with training, so that
the streaming history spans over previous sentences already translated.

\subsection{Partial Bidirectional Encoder}
\label{sec:partial}

In simultaneous MT, the conventional Transformer-based bidirectional
encoder representation (of the $l$-th layer) of a source token at any
position $j$ is constrained to the current $n$-th sentence
\begin{equation}
\label{eqn:encsim}
	e^{(l)}_j = \textrm{Enc}\left(e^{(l-1)}_{a_n:G(i)}\right)
\end{equation} 
where $a_n \leq j \leq G(i)$, while the decoder can only attend to
previous target words and the encoding of those source words that are
available at each timestep
\begin{equation}%
\label{eqn:decsim}
	s^{(l)}_i = \textrm{Dec}\left(s^{(l-1)}_{{b_n}:i-1}, e^{(l-1)}_{{a_n}:G(i)}\right).
\end{equation} 

As a result, the encoder and decoder representations for positions $j$
and $i$, respectively, could be computed taking advantage of
subsequent positions to position $j$ up to position $G(i)$ at
inference time.  However, at training time, this means that this
bidirectional encoding-decoding of the source sentence has to be
computed for every timestep, taking up to $|y|$ times longer than the
conventional Transformer model.

To alleviate this problem, 
\cite{Elbayad2020}~proposes a wait-$k$ simultaneous MT model based 
on a modification of the Transformer architecture that uses
unidirectional encoders and multiple values of $k$ at training time.
In this way, the model is consistent with the limited-input
restriction of simultaneous MT at inference time.  The proposed
unidirectional encoder can be stated as
\begin{equation}
   \label{eqn:elbayad}
	e^{(l)}_j = \textrm{Enc}\left(e^{(l-1)}_{a_n:j}\right),
\end{equation}
that is more restrictive than that in Eq.~\ref{eqn:encsim}, 
and it consequently conditions the decoder representation, 
since $G(i)$ in Eq.~\ref{eqn:decsim} depends on the specific 
$k$ value employed at each training step. 

As mentioned above, the unidirectional encoder just requires a single
forward pass of the encoder at training time, and therefore there is
no additional computational cost compared with a conventional
Transformer. However, it does not take into account all possible input
tokens for different values of $k$. Indeed, the encoding of the $j$-th
input token will not consider those tokens beyond the $j$-th position,
even if including them into the encoding process does not prevent us
from performing a single forward pass.

A trade-off between the unidirectional and bidirectional encoders is
what we have dubbed Partial Bidirectional Encoder (PBE), which
modifies the unidirectional encoder to allow the first $k-1$ source
positions to have access to succeeding tokens according to
\begin{equation}
   \label{eqn:pbe}
	e^{(l)}_j = \textrm{Enc}\left(e^{(l-1)}_{a_n:\max(a_n+k-1,j)}\right).
\end{equation} 

PBE allows for a longer context when encoding the initial positions
and is consistent with Eq.~\ref{eqn:decsim}. At training time a single
forward pass of the encoder-decoder is still possible as in the
unidirectional encoder, and therefore no additional training cost is
incurred. At inference time, we fall back to the bidirectional
encoder.

\begin{figure*}[ht!] 
\centering

\begin{tikzpicture}[scale=0.8, every node/.style={scale=0.8}]
  \draw (1,0) node[draw,circle,label=below:$ X_{1}$,fill=black](x1){};
  \draw (2,0) node[draw,circle,label=below:$ X_{2}$,fill=black](x2){};
  \draw (3,0) node[draw,circle,label=below:$ X_{3}$,fill=black](x3){};
  \draw (4,0) node[draw,circle,label=below:$ X_{4}$,fill=black](x4){};
  \draw (5,0) node[draw,circle,label={below:$ X_{5}$},fill=black](x5){};
  \draw (5,-1) node[draw](G){$G(2)=5$};

  \draw (1,1) node[draw,circle,fill=black](e1){};
  \draw (2,1) node[draw,circle,fill=black](e2){};
  \draw (3,1) node[draw,circle,label=above:$ e_{3}$,fill=black](e3){};
  \draw (4,1) node[draw,circle,fill=black](e4){};
  \draw (5,1) node[draw,circle,fill=black](e5){};

  \draw[->] (x1.north) -- (e3.south);
  \draw[->] (x2.north) -- (e3.south);
  \draw[->] (x3.north) -- (e3.south);
  \draw[->] (x4.north) -- (e3.south);
  \draw[->] (x5.north) -- (e3.south);

  \draw (1,3) node[draw,circle,label=below:$ X_{1}$,fill=black](x1){};
  \draw (2,3) node[draw,circle,label=below:$ X_{2}$,fill=black](x2){};
  \draw (3,3) node[draw,circle,label=below:$ X_{3}$,fill=black](x3){};
  \draw (4,3) node[draw,circle,label=below:$ X_{4}$,fill=black](x4){};
  \draw (5,3) node[draw,circle,label={[gray]below:$ X_{5}$},fill=gray](x5){};
  \draw (5,2) node[draw](G){$G(1)=4$};

  \draw (1,4) node[draw,circle,fill=black](e1){};
  \draw (2,4) node[draw,circle,fill=black](e2){};
  \draw (3,4) node[draw,circle,label=above:$ e_{3}$,fill=black](e3){};
  \draw (4,4) node[draw,circle,fill=black](e4){};
  \draw (5,4) node[draw,circle,fill=gray](e5){};

  \draw[->] (x1.north) -- (e3.south);
  \draw[->] (x2.north) -- (e3.south);
  \draw[->] (x3.north) -- (e3.south);
  \draw[->] (x4.north) -- (e3.south);

\end{tikzpicture} \hspace{5mm}
\begin{tikzpicture}[scale=0.8, every node/.style={scale=0.8}]
  \draw (1,0) node[draw,circle,label=below:$ X_{1}$,fill=black](x1){};
  \draw (2,0) node[draw,circle,label=below:$ X_{2}$,fill=black](x2){};
  \draw (3,0) node[draw,circle,label=below:$ X_{3}$,fill=black](x3){};
  \draw (4,0) node[draw,circle,label=below:$ X_{4}$,fill=black](x4){};
  \draw (5,0) node[draw,circle,label={below:$ X_{5}$},fill=black](x5){};
  \draw (5,-1) node[draw](G){$G(2)=5$};

  \draw (1,1) node[draw,circle,fill=black](e1){};
  \draw (2,1) node[draw,circle,fill=black](e2){};
  \draw (3,1) node[draw,circle,label=above:$ e_{3}$,fill=black](e3){};
  \draw (4,1) node[draw,circle,fill=black](e4){};
  \draw (5,1) node[draw,circle,fill=black](e5){};

  \draw[->] (x1.north) -- (e3.south);
  \draw[->] (x2.north) -- (e3.south);
  \draw[->] (x3.north) -- (e3.south);

  \draw (1,3) node[draw,circle,label=below:$ X_{1}$,fill=black](x1){};
  \draw (2,3) node[draw,circle,label=below:$ X_{2}$,fill=black](x2){};
  \draw (3,3) node[draw,circle,label=below:$ X_{3}$,fill=black](x3){};
  \draw (4,3) node[draw,circle,label=below:$ X_{4}$,fill=black](x4){};
  \draw (5,3) node[draw,circle,label={[gray]below:$ X_{5}$},fill=gray](x5){};
  \draw (5,2) node[draw](G){$G(1)=4$};

  \draw (1,4) node[draw,circle,fill=black](e1){};
  \draw (2,4) node[draw,circle,fill=black](e2){};
  \draw (3,4) node[draw,circle,label=above:$ e_{3}$,fill=black](e3){};
  \draw (4,4) node[draw,circle,fill=black](e4){};
  \draw (5,4) node[draw,circle,fill=gray](e5){};

  \draw[->] (x1.north) -- (e3.south);
  \draw[->] (x2.north) -- (e3.south);
  \draw[->] (x3.north) -- (e3.south);

\end{tikzpicture} \hspace{5mm}
\begin{tikzpicture}[scale=0.8, every node/.style={scale=0.8}]
  \draw (1,0) node[draw,circle,label=below:$ X_{1}$,fill=black](x1){};
  \draw (2,0) node[draw,circle,label=below:$ X_{2}$,fill=black](x2){};
  \draw (3,0) node[draw,circle,label=below:$ X_{3}$,fill=black](x3){};
  \draw (4,0) node[draw,circle,label=below:$ X_{4}$,fill=black](x4){};
  \draw (5,0) node[draw,circle,label={below:$ X_{5}$},fill=black](x5){};
  \draw (5,-1) node[draw](G){$G(2)=5$};

  \draw (1,1) node[draw,circle,fill=black](e1){};
  \draw (2,1) node[draw,circle,fill=black](e2){};
  \draw (3,1) node[draw,circle,label=above:$ e_{3}$,fill=black](e3){};
  \draw (4,1) node[draw,circle,fill=black](e4){};
  \draw (5,1) node[draw,circle,fill=black](e5){};

  \draw[->] (x1.north) -- (e3.south);
  \draw[->] (x2.north) -- (e3.south);
  \draw[->] (x3.north) -- (e3.south);
  \draw[->] (x4.north) -- (e3.south);

  \draw (1,3) node[draw,circle,label=below:$ X_{1}$,fill=black](x1){};
  \draw (2,3) node[draw,circle,label=below:$ X_{2}$,fill=black](x2){};
  \draw (3,3) node[draw,circle,label=below:$ X_{3}$,fill=black](x3){};
  \draw (4,3) node[draw,circle,label=below:$ X_{4}$,fill=black](x4){};
  \draw (5,3) node[draw,circle,label={[gray]below:$ X_{5}$},fill=gray](x5){};
  \draw (5,2) node[draw](G){$G(1)=4$};

  \draw (1,4) node[draw,circle,fill=black](e1){};
  \draw (2,4) node[draw,circle,fill=black](e2){};
  \draw (3,4) node[draw,circle,label=above:$ e_{3}$,fill=black](e3){};
  \draw (4,4) node[draw,circle,fill=black](e4){};
  \draw (5,4) node[draw,circle,fill=gray](e5){};

  \draw[->] (x1.north) -- (e3.south);
  \draw[->] (x2.north) -- (e3.south);
  \draw[->] (x3.north) -- (e3.south);
  \draw[->] (x4.north) -- (e3.south);
\end{tikzpicture}

\caption{
Comparison of attention positions in $j=3$ for bidirectional (left),
unidirectional (center) and PBE (right) encoders with $k=4$ in two
consecutive timesteps $i=1$ with $G(1)=4$ (top) and $i=2$ with
$G(2)=5$ (bottom).
\label{fig:att_example}}
\end{figure*}
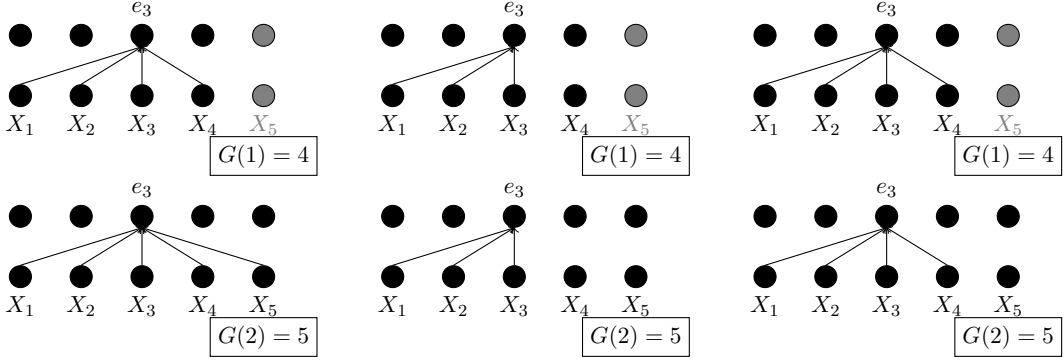

Figure~\ref{fig:att_example} shows a graphical comparison of the
attention mechanism in $j=3$ across the bidirectional (left),
unidirectional (center) and PBE (right) encoders with $k=4$ for two
consecutive timesteps $i=1$ with $G(1)=4$ (top) and $i=2$ with
$G(2)=5$ (bottom). As observed, PBE can take advantage of additional
positions from $j+1$ up to $k$ with respect to the unidirectional
encoder.

In a streaming setup, the bidirectional encoder-decoder of
Eqs.~\ref{eqn:encsim} and~\ref{eqn:decsim} are not necessarily
constrained to the current sentence and could exploit a streaming
history of $H(i)$ tokens
\begin{align}
\label{eqn:encstr}
	e^{(l)}_j &\!=\! \textrm{Enc}\!\left(e^{(l-1)}_{G(i)-H(i)+1:G(i)}\right)\\
\label{eqn:decstr}
	s^{(l)}_i &\!=\! \textrm{Dec}\!\left(s^{(l-1)}_{i-H(i):i-1}, e^{(l-1)}_{G(i)-H(i)+1:G(i)}\right).
\end{align} 

Likewise, the proposed PBE with streaming history states as follows
\begin{equation}
	e^{(l)}_j \!\!\!=\!\! \textrm{Enc} \!\!\left(\!e^{(l-1)}_{G(i)-H(i)+1:\max(G(i)-H(i)+k,j)}\right)\!.
\end{equation}

%\begin{equation*}\label{eqn:encstr}
%	e^{(l)}_j = \textrm{Enc}(e^{(l-1)}_{G(i)-H(i)+1:G(i)})
%\end{equation*}

%\begin{equation*}
%\label{eqn:decstr}
%	s^{(l)}_i = \textrm{Dec}(s^{(l-1)}_{i-H(i):i-1}, e^{(l-1)}_{G(i)-H(i)+1:G(i)})
%\end{equation*}
%
%Likewise, the proposed PBE with streaming history states as follows
%\begin{equation*}
%	e^{(l)}_j = \textrm{Enc} (e^{(l-1)}_{G(i)-H(i)+1:\max(G(i)-H(i)+k,j)})
%\end{equation*}

% \subsection{Decoding using Streaming History}
% 
% \begin{algorithm}
% \caption{Decoding Algorithm for Stream Translation using history}\label{alg:history_decoding}
% \begin{algorithmic}
% \Require Translation model m, $N$ sentences
% \For{$n$ in $1:N$}
% 	\State $H_s \gets \textrm{CLIP}(H_s) $
% 	\State $H_t \gets \textrm{CLIP}(H_t) $
% 	\State $x_n \gets$ \{\}
% 	\State $\hat y_n \gets$ \{\}
% 	\Repeat
% 		\State action $\gets$ policy$_m$($H_s,H_t,x_n,\hat y_n$)
% 		\If{action = READ}
% 			\State $x_n \gets x_n \frown READ(n)$
% 		\Else
% 			\State enc$\; \gets \; $m.encode($[H_s \; x_n]$)
% 			\State tok$\; \gets \; $m.decode(enc,$[H_t \; \hat y_n] $)
% 			\State $\hat y_n \gets \hat y_n \frown $ tok
% 		\EndIf
% 	\Until{tok=EOS}
% 	\State $H_s \gets H_s \frown x_n$
% 	\State $H_t \gets H_t \frown \hat y_n$
% 
% \EndFor
% 
% 
% \end{algorithmic}
% \end{algorithm}

\section{Experimental setup}
\label{sec:exp}

\begin{table}[h!]
\centering
 \caption{Basic statistics of the training data from the 
 IWSLT 2020 Evaluation Campaign (M = Millions). \label{tab:mt_corpus}}
\small 
\begin{tabular}{lcrrr}
	Corpus & Doc          & Sents(M) & \multicolumn{2}{c}{Tokens(M)} \\
	       &              &          & German & English\\\hline
News-Comm. & $\checkmark$ &  0.3     &   7.4  &  7.2   \\ 
Wikititles &              &  1.3     &   2.7  &  3.1   \\ 
Europarl   & $\checkmark$ &  1.8     &  42.5  &  45.5  \\ 
Rapid      & $\checkmark$ &  1.5     &  26.0  &  26.9  \\ 
MuST-C     & $\checkmark$ &  0.2     &   3.9  &  4.2   \\ 
TED        & $\checkmark$ &  0.2     &   3.3  &  3.6   \\ 
LibriVox   &              &  0.1     &   0.9  &  1.1   \\ 
Paracrawl  &              & 31.4     & 465.2  &  502.9 \\
	\end{tabular}
\end{table}

A series of comparative experiments in terms of translation quality
and latency have been carried out using data from the IWSLT 2020
Evaluation Campaign~\cite{Ansari2020}, for both German$\rightarrow$English and
English$\rightarrow$German. For the streaming condition, our
system is tuned on the 2010 dev set, and evaluated on the 2010 test set 
for comparison with \cite{Schneider2020}. Under this setting,
words were lowercased and punctuation was removed in
order to simulate a basic upstream ASR system. Also, a second
non-streaming setting is used for the English$\rightarrow$German direction
to compare our system with top-of-the-line sentence-based
simultaneous MT systems participating in the IWSLT 2020 Simultaneous Translation
Task.

Table~\ref{tab:mt_corpus} summarizes the basic statistics of the IWSLT
corpora used for training the streaming MT systems. Corpora for which
document information is readily available are processed for training
using the sliding window technique mentioned in
Section~\ref{sec:history}. Specifically, for each training sentence,
we prepend previous sentences, which are added one
by one until a threshold $h$ of history tokens is reached. Sentence
boundaries are defined on the presence of special tokens (
\verb <DOC>, \verb <CONT>, \verb <BRK>, \verb <SEP> ) as
in~\cite{junczys2019microsoft}. Byte Pair 
Encoding~\cite{sennrich2016neural} with 40K merge operations is applied 
to the data after preprocessing.

Our streaming MT system is evaluated in terms of latency and
translation quality with BLEU~\cite{Papineni2002}.  Traditionally,
latency evaluation in simultaneous MT has been carried out using 
AP, AL and DAL. However, these measures have been devised
for sentence-level evaluation, where the latency of every sentence is
computed independently from each other and as mentioned before, 
they do not perform well on
a streaming setup. Thus, we revert to the stream-based adaptation of
these measures proposed in~\cite{Iranzo2021stream} unless stated
otherwise.

Latency measures for a sentence pair $(\bm{x},\bm{y})$ are based on a
cost function $C_i(\bm{x},\bm{y})$ and a normalization term
$Z(\bm{x},\bm{y})$
\begin{equation}
\label{eqn:L}
L(\bm{x},\bm{y}) = \frac{1}{Z(\bm{x},\bm{y})} \sum_{i} C_i(\bm{x},\bm{y})
\end{equation}
where 
 
\begin{equation}
\label{eqn:C}
  C_i(\bm{x},\bm{y}) = 
  \begin{cases}
    g(i) & \text{AP}\\
    g(i) - \frac{i-1}{\gamma} & \text{AL}\\
    g'(i) - \frac{i-1}{\gamma} & \text{DAL}
  \end{cases} 
\end{equation}
and 
\begin{equation}
\label{eqn:Z}
  Z(\bm{x},\bm{y}) = 
  \left\{ \!\!\! \begin{array}{ll}
    |\bm{x}|\cdot|\bm{y}| & \text{AP}\\
    \argmin\limits_{i:g(i)=|\bm{x}|} \; i & \text{AL}\\
    |\bm{y}| & \text{DAL}
  \end{array} \right. 
\end{equation}

Latency measures can be computed in a streaming manner by considering
a global delay function $G(i)$, that is mapped into a relative delay
so that it can be compared with the sentence-level oracle delay. For
the $i$-th target position of the $n$-th sentence, the associated
relative delay can be obtained from the global delay function as
$g_n(i)=G(i+b_n)- a_n$. So, the stream-adapted cost function of
the latency measures is defined as
 
\begin{equation}
\label{eqn:sC}
  C_{i}(\bm{x}_n,\bm{y}_n) = 
  \begin{cases}  
    g_n(i) & \text{AP}\\
    g_n(i) - \frac{i-1}{\gamma_n} & \text{AL}\\
    g_n'(i) - \frac{i-1}{\gamma_n} & \text{DAL}\\
  \end{cases} 
\end{equation}
with $g_n'(i)$ defined as
% 
% {\small
\begin{equation}
\max 
\begin{cases}
g_n(i)\\
%\begin{cases}
 \!\!\left\{ \!\!\!\! \begin{array}{l@{~~}l}
g_{n-1}'(|\bm{x}_{n-1}|) + \frac{1}{\gamma_{n-1}} & i=1\\
g_n'(i-1) + \frac{1}{\gamma_{n}} & i>1
 \end{array} \right. 
%\end{cases} 
\end{cases} 
\end{equation}

This definition assumes that the source and target sentence
segmentation of the stream are uncovered, but this is not always the
case~\cite{Schneider2020} or they may not match that of the reference
translations. However, sentence boundaries can be obtained by
re-segmenting the system hypothesis following exactly the same
procedure applied to compute translation quality in ST evaluation. To
this purpose, we use the MWER segmenter~\cite{MatusovLBN05} to compute
sentence boundaries according to the reference translations.

Our streaming MT models have been trained following the conventional
Transformer BASE (German$\leftrightarrow$English streaming MT) and BIG 
(English$\rightarrow$German simultaneous MT) configurations~\cite{VaswaniSPUJGKP17}.
As in~\cite{Schneider2020}, after training is finished, the models are
finetuned on the training set of MuST-C~\cite{Gangi2019d}.

The proposed model in Section~\ref{sec:streaming} assumes that at
inference time the source stream has been segmented into sentences.
To this purpose, we opt for the text-based DS
model~\cite{Iranzo2020b}, a sliding-window segmenter that moves over
the source stream taking a split decision at each token based on a
local-context window that extends to both past and future tokens.
This segmenter is streaming-ready and obtains superior translation
quality when compared with other segmenters~\cite{Stolcke2002,Cho2017}.
As the future window length of the DS segmenter conditions the latency
of the streaming MT system, this length was adjusted to find a tradeoff
between latency and translation quality.  The DS segmenter was trained
on the TED corpus~\cite{Cettolo2012}.

\section{Evaluation}
\label{sec:eva}

% \subsection{German $\rightarrow$ English}

\begin{figure}[h!] 
\centering
\includegraphics[width=.42\textwidth]{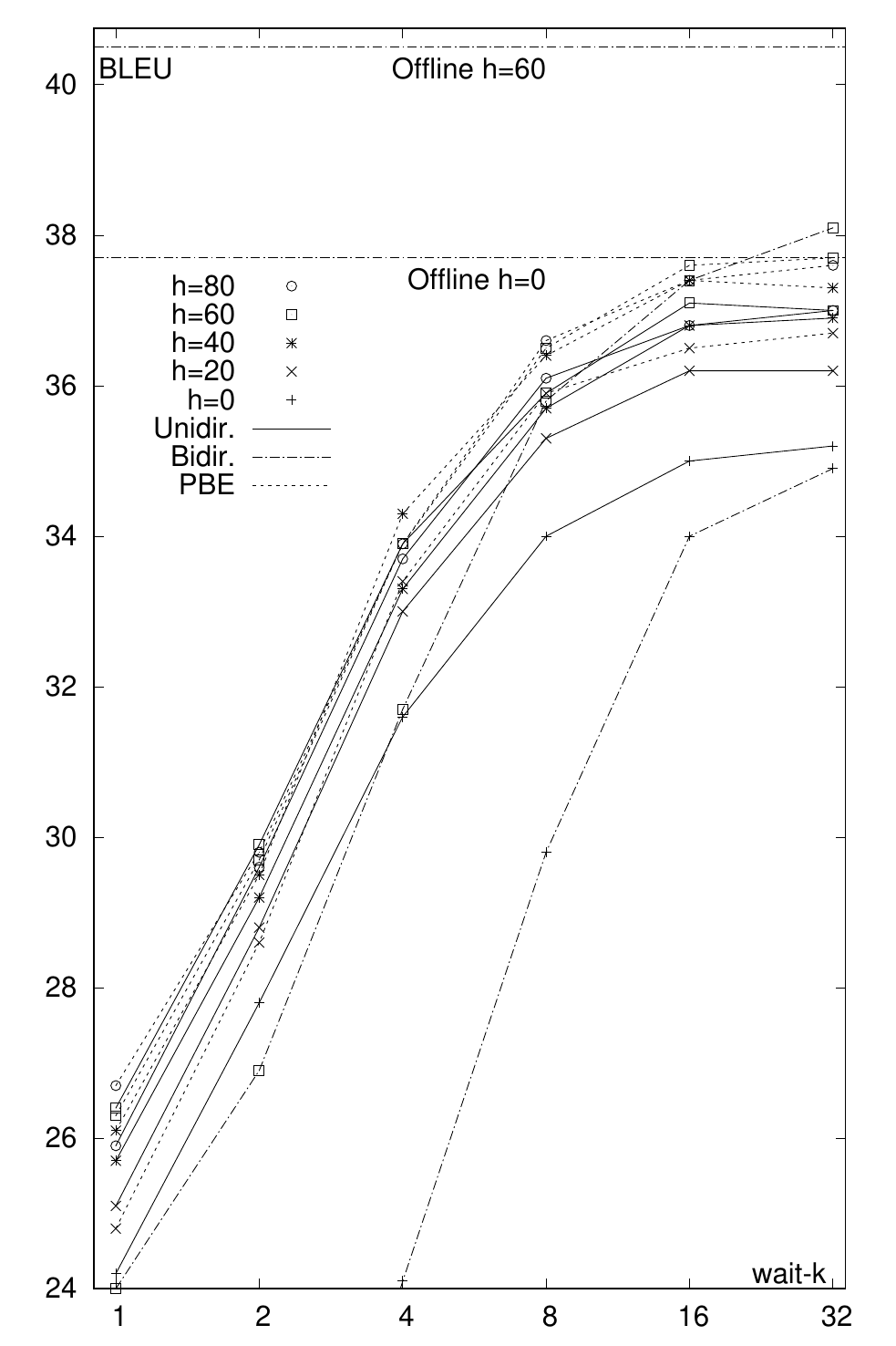}
\caption{
BLEU scores on the German-English IWSLT 2010 dev set as a function of
the $k$ value in the wait-$k$ policy for a range of streaming history
($h)$ lengths and encoder type (See Appendix \ref{app:extra_results} for a close-up).
\label{fig:reference_comparison}}
\end{figure}

Figure~\ref{fig:reference_comparison} reports the evolution of BLEU
scores on the German-English IWSLT 2010 dev set as a function of the
$k$ value in the wait-$k$ policy for a range of streaming history
lengths ($h=\{0,20,40,60,80\}$). We show results for the 3 encoders introduced previously. History lengths were selected taking into
account that the average sentence length is 20 tokens. A history length of
zero ($h=0$) refers to the conventional sentence-level simultaneous MT
model. The BLEU scores for the offline MT systems with a bidirectional 
encoder are also reported using horizontal lines, in order to serve
as reference values. We report offline results for $h=0$ and the best performing history
configuration, $h=60$. All systems used the reference segmentation during decoding.

\begin{figure*}[h!] 
\includegraphics[width=.5\textwidth]{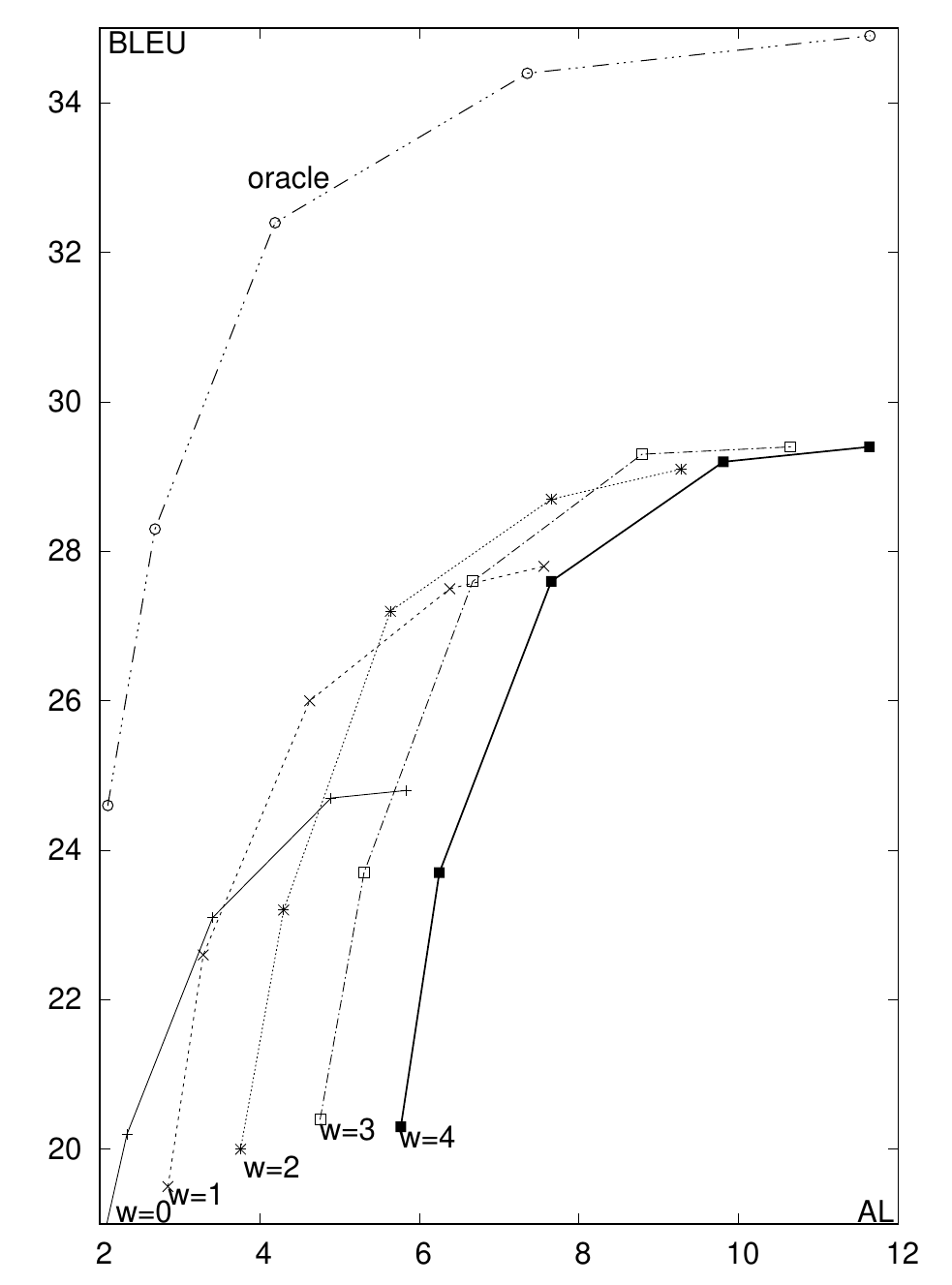}%
\includegraphics[width=.5\textwidth]{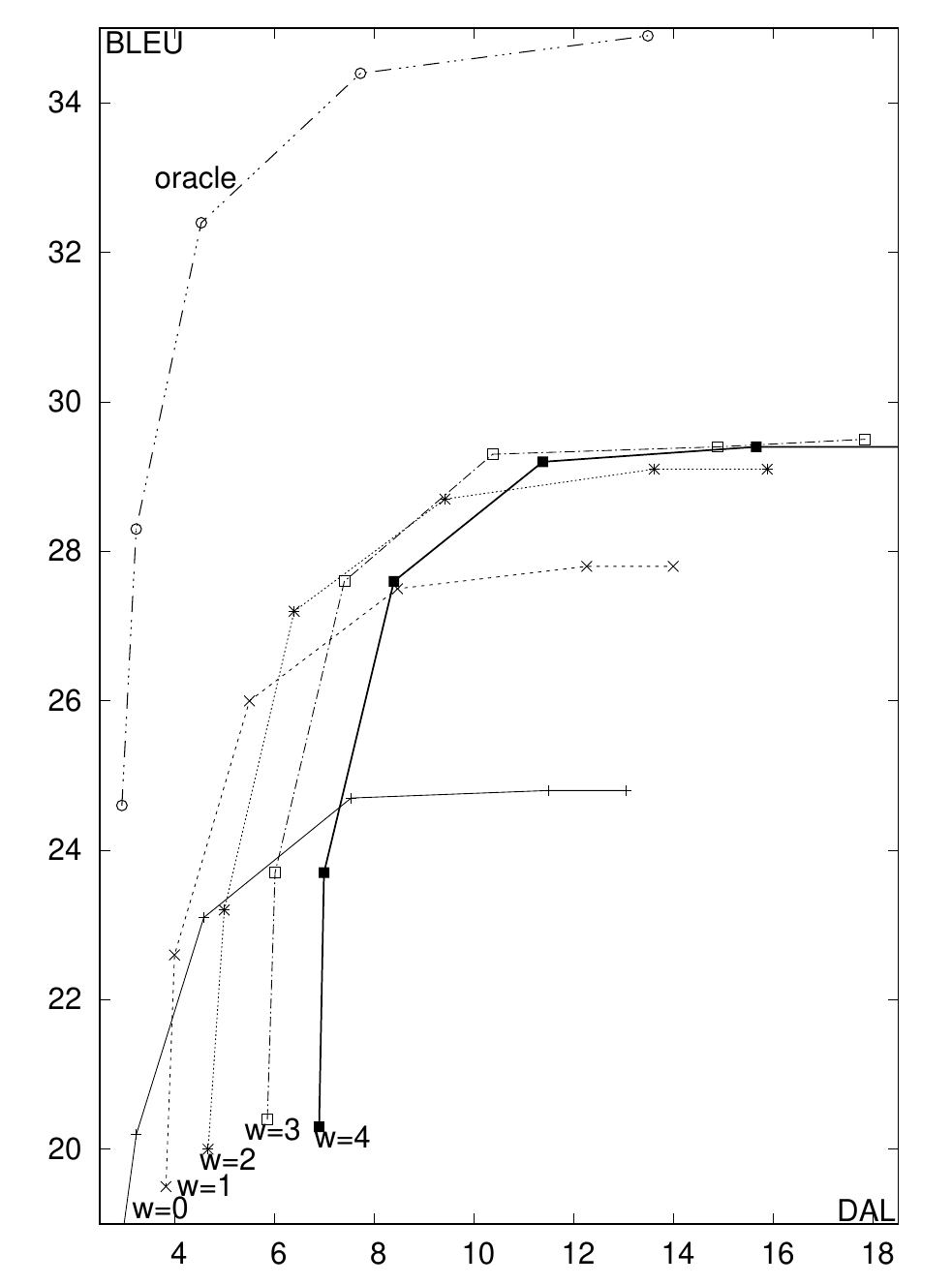}
\caption{
BLEU scores versus stream-adapted AL and DAL (scale $s$=0.85) with
segmenters of future window length $w=\{0,1,2,3,4\}$ on the IWSLT 2010
test set. Points over each curve correspond to $k=\{1,2,4,8,16\}$
values of the wait-$k$ policy used at inference
time.\label{fig:latency_streaming}}
\end{figure*}

As observed, BLEU scores of the simultaneous MT systems leveraging on
the streaming history ($h>0$) are systematically and notably higher
than those of conventional sentence-based simultaneous MT system
($h=0$) over the range of wait-$k$ values. Indeed, as the streaming
history increases, BLEU scores also do reaching what it seems the
optimal history length at $h=60$ and slightly degrading at $h=80$.  As
expected, when replacing the unidirectional encoder by the PBE, BLEU
scores improve as the wait-$k$ value increases, since PBE
has additional access to those tokens from $j+1$ up to $k$.
For instance, for $k=32$ and $h=60$, PBE is $0.7$ BLEU points 
above the unidirectional encoder.
%      Uni  PBE
% k=8  35.9 36.3
% k=16 36.7 37.2
% k=32 36.7 37.2
%In fact, the BLEU
%score of this latter system converges to the offline bidirectional
%system for $k=32$, though we should keep in mind that training
%procedures of the streaming MT system (prefix training) and that of
%the offline bidirectional system (sentence based) are different.
On the other hand, it can be observed how using an encoder which is not fully bidirectional
during training, creates a performance gap with respect to the offline bidirectional model
when carrying out inference in an offline manner ($k \geq 32$). 
It can 
be also observed how the PBE model is better prepared for this scenario and shows a smaller gap. It is important to keep in mind
that although both offline and PBE models behave the same way during inference for a large enough $k$, during training 
time the PBE model, trained using the multi-$k$ with $k$ randomly sampled for each batch, has been optimized jointly for low, medium and high latencies.

%Decir algo de explicit latencies?

In general, the bidirectional encoder shows poor performance for simultaneous MT. This can be explained by the fact that there exists a mismatch between the training condition (whole source  available) and the inference condition (only a prefix of the source is available for $k<32$). These results are consistent with \cite{Elbayad2020}. Keep in mind that this bidirectional model is different from the offline one because it has been subject to the constraints of Eq. \ref{eqn:decsim} during training. As a result of the BLEU scores reported in
Figure~\ref{fig:reference_comparison}, the streaming MT system with
$h=60$ and PBE was used in the rest of the German-English experiments.

Following~\cite{Schneider2020}'s setup, the test set is lowercased and
concatenated into a single stream.  In order to measure the latency of
the pipeline defined by the segmenter followed by MT system, it is
necessary to take into account not only the latency of the MT system
but also that of the segmenter. Thankfully this is straightforward to
do in our pipeline, as a segmenter with a future window of length $w$
modifies the pipeline policy so that, at the start of the stream, $w$
READ actions are carried out to fill up the future window. Then, every
time the MT system carries out a READ action, it receives one
token from the segmenter. Thus, the integration of the segmenter into
the pipeline is transparent from a latency viewpoint.
Figure~\ref{fig:latency_streaming} shows BLEU scores versus
stream-adapted AL and DAL ($s$ scale = 0.85) figures reported with
segmenters of future window length $w=\{0,1,2,3,4\}$ for a streaming
evaluation on the IWSLT 2010 test set. Points over each curve
correspond to $k=\{1,2,4,8,16\}$ values of the wait-$k$ policy used at
inference time. Results for a $w=0$ oracle are also shown as an upper-bound.

As shown, stream-adapted AL and DAL figures achieved by our streaming
MT system are reasonable, lagging 2-10 tokens behind the speaker 
for nearly maximum BLEU scores with
a best BLEU score of 29.5 points.  The same happens with AP figures
ranging from 0.6 for $w=0$ to 1.3 for $w=4$. These figures highlight
the advantages of tying together our translation policy with the
sentence segmentation provided by the DS model.  Every time the DS
model emits an end-of-sentence event, the MT model is forced to
catch-up and translate the entire input. In this way, the MT model
never strays too far from the speaker, even if the source-target
length ratio differs from the $\gamma$ defined at inference time.
See Appendix \ref{app:extra_results} for streaming translation results in the
reverse direction (English $\rightarrow$ German).

\begin{table}[b!]
\centering
 \caption{Latency and quality comparison of ACT ~\cite{Schneider2020} and the proposed STR-MT on the IWSLT 2010 De-En test set.  \label{tab:act_comparison}}

\begin{tabular}{l||rrrr}
Model & BLEU & AP & AL &DAL \\ \hline
ACT & 30.3 & 10.3 & 100.1 & 101.8 \\
STR-MT & 29.5 & 1.2 & 11.2 & 17.8 \\
\end{tabular}

\end{table}

Next, we compare our proposed streaming MT (STR-MT) model with the $\lambda=0.3$ ACT
system~\cite{Schneider2020} in terms of BLEU score and stream-adapted
latency measures on Table \ref{tab:act_comparison}. Stream-level AL and DAL
indicate that the ACT models lags around 100 tokens behind the
speaker. Although both MT systems achieve similar translation quality
levels, they do so at significantly different latencies, since the ACT
model lacks a catch-up mechanism to synchronize and keep the pace of
the speaker.

%This system achieves 30.3 BLEU points and, 10.3,
%100.1 and 101.8 for AP, AL and DAL, respectively.

% Due to the
% normalization term for AP being dependent on $|x_n|$, an AP of 10.3
% must be interpreted as the model lagging behind by $10.3|x_n|$ tokens,
% on average. 
%It is easier to interpret the behaviour of the model using stream-level AL and DAL,
%which indicate that the ACT models lags around 100 tokens behind the
%speaker. Although both MT systems achieve similar translation quality
%levels, they do so at significantly different latencies, since the ACT
%model lacks a catch-up mechanism to syncronize and keep the pace of
%the speaker.

% \subsection{English $\rightarrow$ German}

\begin{figure}[h]
\centering
\includegraphics[width=.5\textwidth]{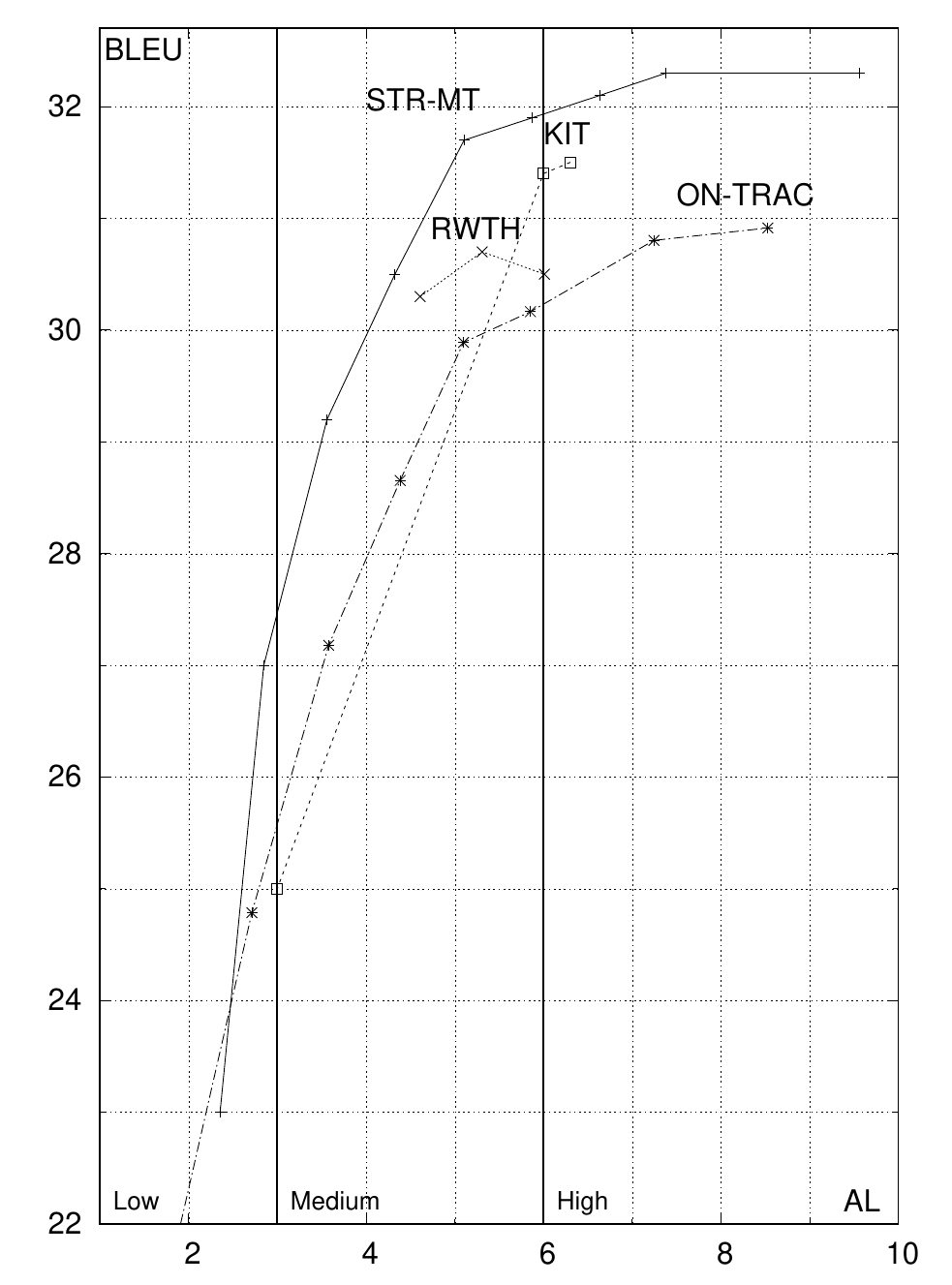}
\caption{
Comparative BLEU scores versus AL at three regimes, low, medium, and high latency, for IWSLT 2020 simultaneous text-to-text track
participants, RWTH, ON-TRAC, KIT and our streaming MT (STR-MT) system 
on the MuST-C corpus.
\label{fig:iwslt-eval}}
\end{figure}

The STR-MT model is now compared 
on the English-German  
IWSLT 2020 simultaneous text-to-text track~\cite{Ansari2020} 
with other participants: RWTH~\cite{bahar-etal-2020-start}, 
KIT~\cite{pham-etal-2020-kits} and ON-TRAC~\cite{elbayad-etal-2020-trac}.
This comparison is carried out in order to assess whether the proposed
streaming MT system is competitive with highly optimized systems for a
simultaneous MT task.  Given that the test set of this track remains
blind, we use the results reported on the MuST-C corpus as a
reference. In order to evaluate all systems under the same conditions,
the reference segmentation of the MuST-C corpus is used instead of the
DS model. Additionally, given that all other participants translate
each sentence independently, the conventional sentence-level AL
latency measure is reported. Figure~\ref{fig:iwslt-eval} shows the
comparison of BLEU scores versus AL measured in terms of detokenized
tokens. As defined in the IWSLT text-to-text track, three AL regimes, low
($\text{AL} \leq 3$), medium ($3 < \text{AL} \leq 6$) and high ($6
< \text{AL} \leq 15$) were considered. %As in 
%Figure~\ref{fig:latency_streaming}, the trade-off
%between translation quality and latency is controlled by the $k$ values
%of the wait-$k$ policy at inference time.

%ON-TRAC and our streaming MT system exhibit similar performance in low
%and medium latency regimes, which is to be expected given that they are
%both based on the multi-$k$ approach. However, at higher latencies,
%our system performs much better thanks to the streaming history and the 
%PBE. This is
%consistent with the results of Figure~\ref{fig:reference_comparison},
%that shows that BLEU improvements achieved by our system 
%increase with the $k$ value of the wait-$k$ policy.

ON-TRAC and our streaming MT system exhibit a similar progression,
which is to be expected given that they are
both based on the multi-$k$ approach. However,
our system consistently outperforms the ON-TRAC system by 1-2 BLEU.
This confirms the importance of utilizing streaming history in order to
significantly improve results, and how the proposed PBE
model can take better advantage of the history.

RWTH and KIT systems are closer in translation quality to our proposal 
than ON-TRAC, for AL between 5 and 7. However, these systems do not show a flexible 
latency policy and are not comparable to our system at other  
regimes. Indeed, for that to be possible, these systems need to be 
re-trained, in contrast to our system in which latency is adjusted at 
inference time.

\section{Conclusions}
\label{sec:con}

In this work, a formalization of streaming MT as a generalization of
simultaneous MT has been proposed in order to define a theoretical
framework in which our two contributions have been made. On the one
hand, we successfully leverage streaming history across sentence
boundaries for a simultaneous MT system based on multiple wait-k paths
that allows our system to greatly improve the results of the
sentence-level baseline. On the other hand, our PBE is able to take
into account longer context information than its unidirectional
counterpart, while keeping the same training efficiency.

Our proposed MT system has been evaluated under a realistic streaming
setting being able to reach similar translation quality than a
state-of-the-art segmentation-free streaming MT system at a fraction
of its latency. Additionally, our system has been shown to be
competitive when compared with state-of-the-art
simultaneous MT systems optimized for sentence-level translation,
obtaining excellent results using a single model across a wide range
of latency levels, thanks to its flexible inference policy.

In terms of future work, additional training and inference procedures
that take advantage of the streaming history in streaming MT are still
open for research. One important avenue of improvement is to devise more robust
training methods, so that simultaneous models can perform as well as their
offline counterparts when carrying out inference
at higher latencies. The segmentation model, though proved useful in a
%streaming setup, adds complexity and can greatly affect translation
%quality. Thus, the development of segmentation-free streaming MT
%models with latency constraints, it is another interesting
%research topic to be addressed.
streaming setup, adds complexity and can greatly affect translation
quality. Thus, the development of segmentation-free streaming MT
models is another interesting research topic.

\section*{Acknowledgements}
The research leading to these results has received funding from the European
Union's Horizon 2020 research and innovation programme under grant agreements
no. 761758 (X5Gon) and 952215 (TAILOR), and Erasmus+ Education programme
under grant agreement no. 20-226-093604-SCH (EXPERT); the Government of
Spain's grant RTI2018-094879-B-I00 (Multisub) funded by
MCIN/AEI/10.13039/501100011033 \& ``ERDF A way of making Europe'', and FPU
scholarships FPU18/04135; and the Generalitat Valenciana's
research project Classroom Activity Recognition (ref. PROMETEO/2019/111). The authors gratefully acknowledge the computer resources at Artemisa, funded by the European Union ERDF and Comunitat Valenciana as well as the technical support provided by the Instituto de Física Corpuscular, IFIC (CSIC-UV).

%\clearpage

\bibliography{anthology,custom}
\bibliographystyle{acl_natbib}

\appendix

\section{Extended Streaming Translation Results}\label{app:extra_results}

\begin{figure}[h!] 
\centering
\includegraphics[width=.42\textwidth]{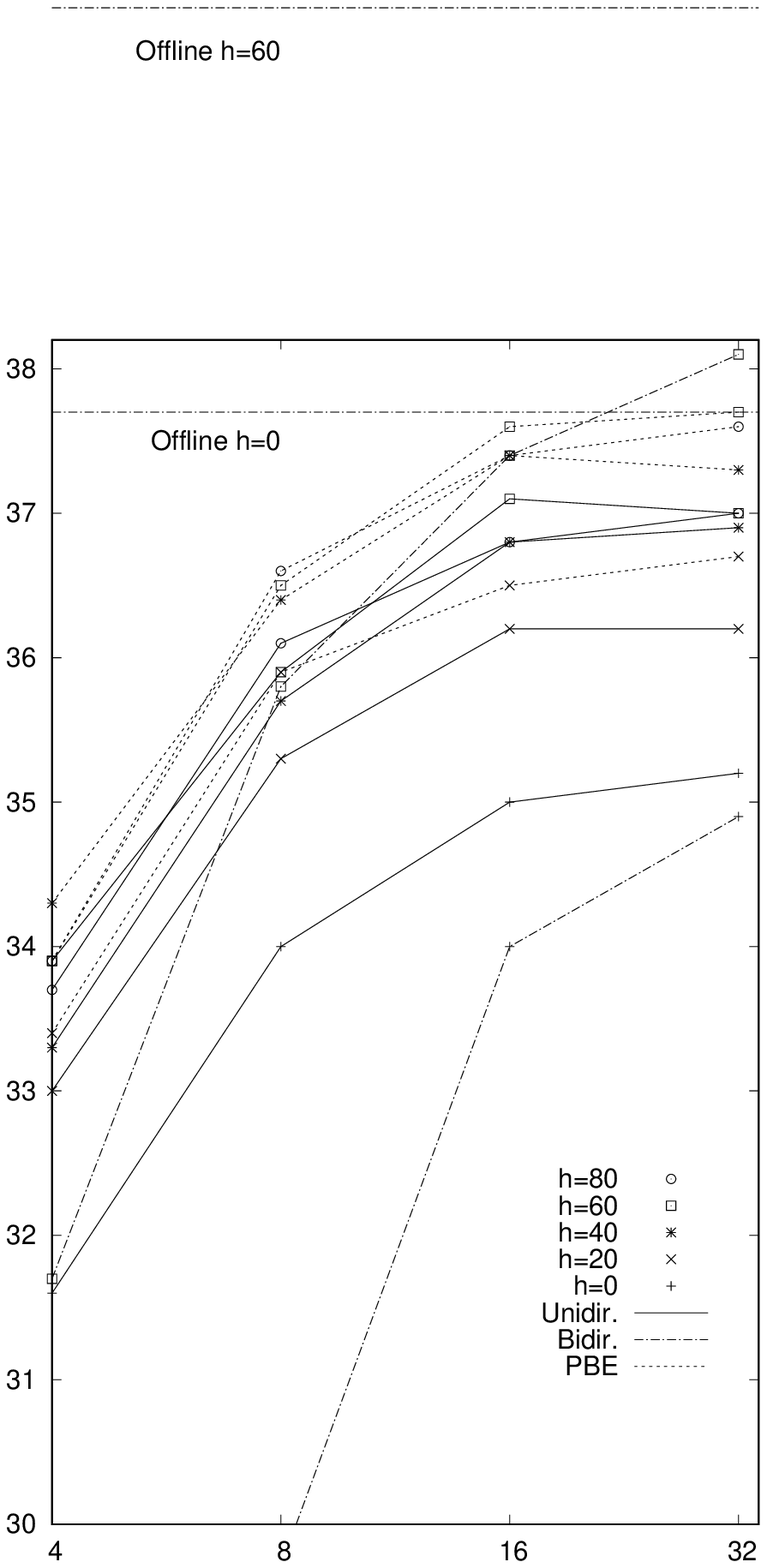}
\caption{BLEU scores on the German-English IWSLT 2010 dev set as a function of
the $k$ value in the wait-$k$ policy for a range of streaming history
($h)$ lengths with a unidirectional encoder (solid lines), PBE (dashed
line) or bidirectional (dashed line with points). This is a close-up
of Figure \ref{fig:reference_comparison}.
\label{fig:reference_comparison_zoomed}}
\end{figure}

Figure \ref{fig:reference_comparison_zoomed} shows a close-up of
Figure \ref{fig:reference_comparison}, which contains results for the
German-English IWSLT 2010 dev set. We can observe how the PBE models
obtain consistent quality improvements over their unidirectional
counterparts.

\begin{figure}[h!] 
\centering
\includegraphics[width=.42\textwidth]{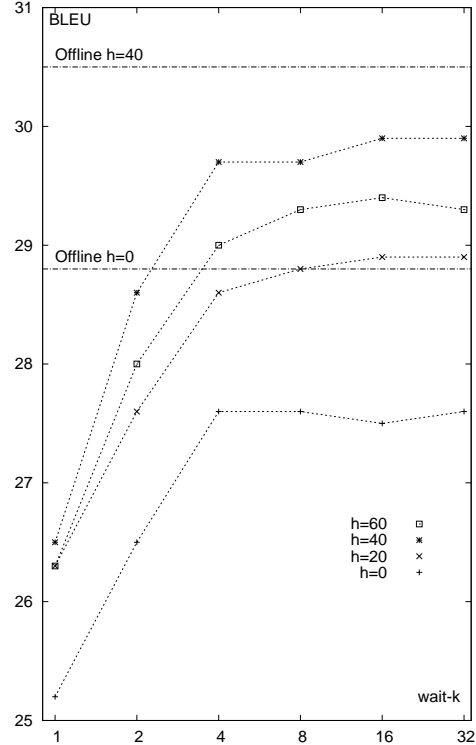}
\caption{
BLEU scores on the English-German IWSLT 2010 dev set as a function of
the $k$ value in the wait-$k$ policy for a range of streaming history
($h)$ lengths using a PBE encoder.
\label{fig:reference_comparison_ende}}
\end{figure}

Apart from the previously reported German $\rightarrow$ English
streaming MT results, we have also conducted experiments in the
reverse direction, English $\rightarrow$ German. These are shown in
Figure \ref{fig:reference_comparison_ende}. The results show a similar
trend to previous experiments, with the addition of streaming
history allowing our systems to obtain significant improvements over
the sentence-based baseline. Unlike the previous case, the optimum
history size in this case is $h=40$ instead of $h=60$.

\begin{figure*}[h!] 
\includegraphics[width=.5\textwidth]{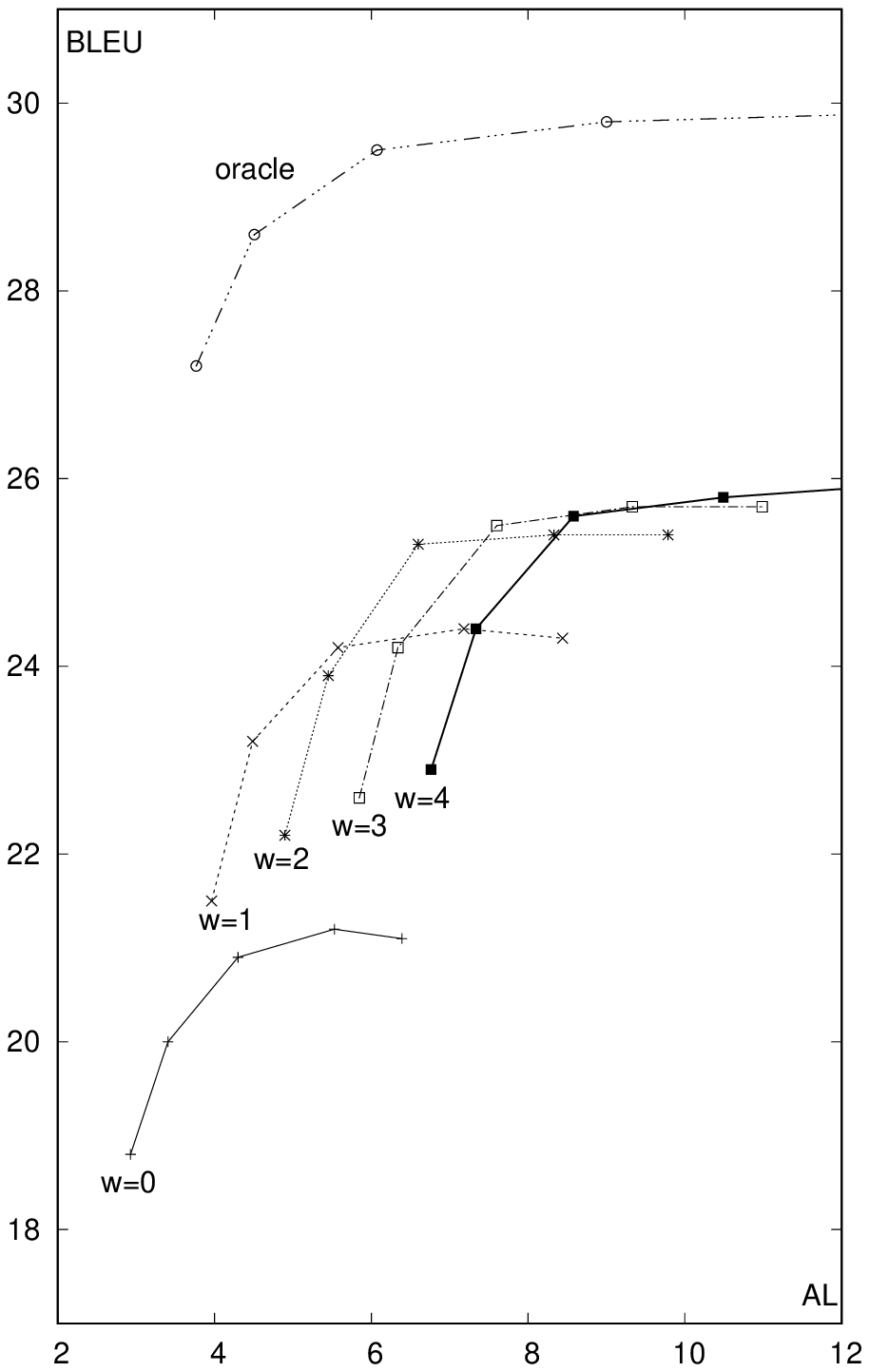}%
\includegraphics[width=.5\textwidth]{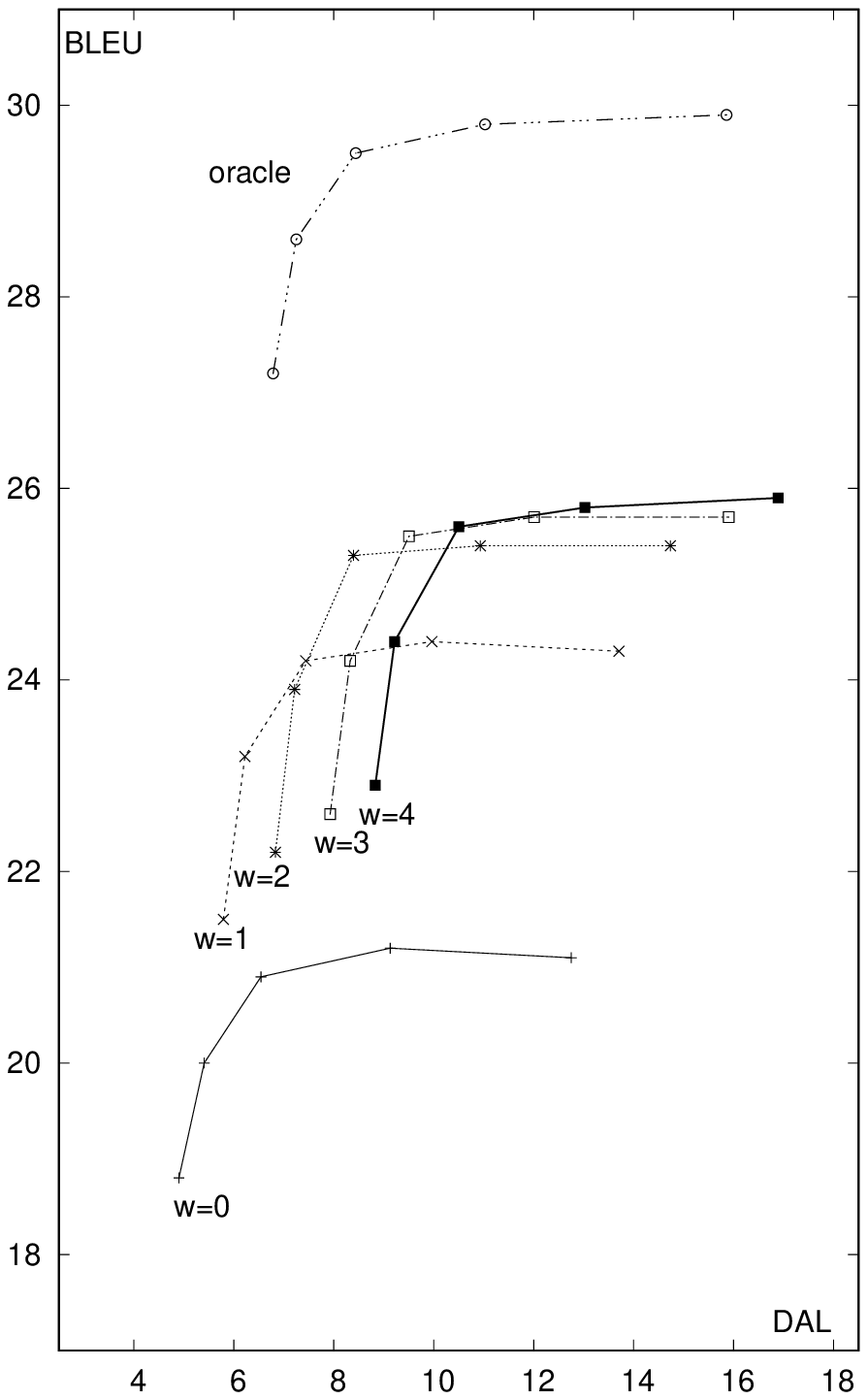}
\caption{
BLEU scores versus stream-adapted AL and DAL (scale $s$=0.85) with
segmenters of future window length $w=\{0,1,2,3,4\}$ on the
English-German IWSLT 2010 test set. Points over each curve correspond
to $k=\{1,2,4,8,16\}$ values of the wait-$k$ policy used at inference
time.\label{fig:latency_streaming_ende}}
\end{figure*}

In order to enable streaming translation, the best performing $h=40$
systems has been combined with a German DS system. Similarly to
previous experiments, we have conducted tests using different values
of $w$ and $k$ in order to balance the latency-quality trade-off,
shown in Figure \ref{fig:latency_streaming_ende}.  Under the streaming
condition, the wait-$k$ policy and DS model allow the model to follow
closely the speaker while achieving good quality, with a latency that
can be easily adjusted between 4 and 15 tokens depending on the
requirements of the task. There are diminishing returns when
increasing the latency above 6-7 tokens, as only marginal gains in
quality are obtained.

\section{Efficiency of the proposed models}
During training of the unidirectional and PBE encoders, the
constraints imposed by Eqs.~\ref{eqn:elbayad} and~\ref{eqn:pbe} are
efficiently implemented by full self-attention,
as in the bidirectional encoder, followed by an attention mask,
for each token to only attend those tokens fulfilling the
constraints. The attention mask sets the weights of the other tokens to
$-\infty$ before application of the self-attention softmax. 
This is exactly
the same mechanism used in the standard Transformer decoder to 
prevent the auto-regressive decoder from accessing future
information.

This means that the three encoder types have
an identical computational behavior. We are not aware of alternative
GPU-based acceleration techniques to speed up the training of the
unidirectional encoder. If so, this could be also applicable to the
training of the standard Transformer decoder.

During inference time, however, the unidirectional encoder has some
advantages. Given that the unidirectional encoder is incremental,
meaning that the encodings of old tokens do not change when a new token
becomes available, the process can be sped up by only computing the 
encoding of
the newly available token. Although encoder self-attention still needs to be
computed, a single vector is used as the query instead of the full
matrix. Table~\ref{tab:effi} shows inference statistics for the
different components of the En $\rightarrow$ De Transformer Big with
h=60. Two setups have been tested: CPU-only inference, and GPU
inference. Results were obtained on an Intel
i9-7920X machine with an NVIDIA GTX 2080Ti.

\begin{table}
\centering
 \caption{Latency of translating a token (in seconds) for the proposed En-De h=60 Transformer Big model. \label{tab:effi}}
\begin{tabular}{lrr}
Component & CPU & GPU \\
\hline
\hline
Unidir. Encoder & 0.034s & 0.002s \\
Bidir. Encoder &  0.138s & 0.002s \\
\hline
Decoder & 0.242s & 0.004s \\
\end{tabular}
\end{table}

The unidirectional encoder is four times faster than the bidirectional
encoder when run on a CPU. However, both encoders perform the same
when run on a GPU. For the streaming MT scenario considered in this work, 
no latency reduction is gained by not re-encoding previous 
tokens due to the GPU paralellization capability. When run on a GPU, 
the proposed model works seamlessly under real-time constraints.

\section{MT System configuration}
The multi-$k$ systems have been trained with the official
implementation (\url{https://github.com/elbayadm/attn2d}). Models are
trained for 0.5M steps on a machine with 4 2080Ti GPUs. Total training
time was 40h for BASE models, and 60h for BIG models. The following
command was used to train them:

\small
\begin{Verbatim}[tabsize=2]
	cri=label_smoothed_cross_entropy;
	ex=simultaneous_translation
	fairseq-train $CORPUS_FOLDER \
	-s $SOURCE_LANG_SUFFIX \
	-t $TARGET_LANG_SUFFIX \
	--user-dir $FAIRSEQ/examples/$ex \
	--arch $ARCH waitk_transformer_base \
	--share-decoder-input-output-embed \
	--left-pad-source False \
	--multi-waitk \
	--optimizer adam \
	--adam-betas '(0.9, 0.98)' \
	--clip-norm 0.0 \
	--lr-scheduler inverse_sqrt \
	--warmup-init-lr 1e-07 \
	--warmup-updates 4000 \
	--lr 0.0005 \
	--min-lr 1e-09 \
	--dropout 0.1 \
	--weight-decay 0.0 \
	--criterion $cri \
	--label-smoothing 0.1 \
	--max-tokens $TOK \
	--update-freq 2 \
	--save-dir $MODEL_OUTPUT_FOLDER \
	--no-progress-bar \
	--log-interval 100 \
	--max-update 500000 \
	--save-interval-updates 10000 \
	--keep-interval-updates 20 \
	--ddp-backend=no_c10d \
	--fp16
\end{Verbatim}
\normalsize

with \begin{verbatim} ARCH=waitk_transformer_base;
 TOK=4000 \end{verbatim} for the BASE configuration,
and \begin{verbatim} ARCH=waitk_transformer_big;
 TOK=2000  \end{verbatim} for the BIG one.

For finetuning, we change to the following:
{ 
\begin{verbatim}
  --lr-scheduler fixed \
  --lr 4.47169e-05 \
\end{verbatim}
}

For the streaming translation scenario, the data is lowercased and all
punctuation signs are removed. For the simultaneous scenario (IWSLT
2020 simultaneous text- to-text), it is truecased and tokenized using
Moses. We apply language identification to the training data using langid
\cite{DBLP:conf/acl/LuiB12} and discard those sentences that have been tagged
with the wrong language. SentencePiece \cite{DBLP:conf/emnlp/KudoR18} is used to
learn the BPE units, and we use whitespace as a suffix in order to know when an entire target word
has been written during decoding.

In order to obtain samples that can be used for training
streaming MT models, a sliding window that moves over whole sentences
is used to extract consistent source-target
samples. Figure \ref{fig:corpus_ex} shows an example of corpus
construction using $h=5$. The generated streaming data is upsampled to keep a 1-to-3 ratio
with the regular sentence-level data.

\begin{figure*}

\centering

\begin{tabular}{lll}
Sentece pair & Source & Target \\
\hline \hline
1& $ x_{1,1} \: x_{1,2} $ & $ y_{1,1} \: y_{1,2}    $ \\ 
2 &$ x_{2,1} \: x_{2,2} \: x_{2,3} $ & $ y_{2,1} \: y_{2,2}    $ \\
3& $ x_{3,1} \: x_{3,2} \: x_{3,3} $ & $ y_{3,1} \: y_{3,2} \: y_{3,3}   $ \\
4 &$ x_{4,1} \: x_{4,2} $ & $ y_{4,1} \: y_{4,2}    $ \\
\end{tabular}

\vspace{10mm}

%\begin{tabular}{ll}
%Source & Target \\
%\hline \hline
%<DOC> $ x_{1,1} \: x_{1,2} $ <BRK> & <DOC> $ y_{1,1} \: y_{1,2} $ <BRK> \\ 
%{\color{gray} <DOC> $ x_{1,1} \: x_{1,2} $} <SEP> $ x_{2,1} \: x_{2,2} \: x_{2,3} $ <BRK> & {\color{gray} <DOC> $ y_{1,1} \: y_{1,2} $} <SEP> $ y_{2,1} \: y_{2,2} $ <BRK> \\ 
%{\color{gray} <DOC> $ x_{1,1} \: x_{1,2} $ <SEP> $ x_{2,1} \: x_{2,2} \: x_{2,3} $} <SEP> $ x_{3,1} \: x_{3,2} \: x_{3,3} $ <BRK> & %{\color{gray} <DOC> $ y_{1,1} \: y_{1,2} $ <SEP> $ y_{2,1} \: y_{2,2} $} <SEP> $ y_{3,1} \: y_{3,2} \: y_{3,3}$ <BRK> \\ 
%{\color{gray}<CONT> $  x_{3,1} \: x_{3,2} \: x_{3,3} $} <SEP> $x_{4,1}  x_{4,2}  $ <END>  & {\color{gray} <CONT> $  y_{3,1} \: y_{3,2} \: y_{3,3} $} <SEP> $y_{4,1}  y_{4,2}$ <END> \\
%\end{tabular}

\begin{tabular}{cl}
Sentence pair & Source \\
\hline \hline
1 & <DOC> $ x_{1,1} \: x_{1,2} $ <BRK>  \\ 
2 & {\color{black!45} <DOC> $ x_{1,1} \: x_{1,2} $} <SEP> $ x_{2,1} \: x_{2,2} \: x_{2,3} $ <BRK>  \\ 
3 & {\color{black!45} <DOC> $ x_{1,1} \: x_{1,2} $ <SEP> $ x_{2,1} \: x_{2,2} \: x_{2,3} $} <SEP> $ x_{3,1} \: x_{3,2} \: x_{3,3} $ <BRK> \\ 
4 &{\color{black!45} <CONT> $  x_{3,1} \: x_{3,2} \: x_{3,3} $} <SEP> $x_{4,1}  x_{4,2}  $ <END>  \\
\\
Sentence pair & Target \\
\hline \hline
1 & <DOC> $ y_{1,1} \: y_{1,2} $ <BRK> \\ 
2 & {\color{black!45} <DOC> $ y_{1,1} \: y_{1,2} $} <SEP> $ y_{2,1} \: y_{2,2} $ <BRK> \\ 
3 & {\color{black!45} <DOC> $ y_{1,1} \: y_{1,2} $ <SEP> $ y_{2,1} \: y_{2,2} $} <SEP> $ y_{3,1} \: y_{3,2} \: y_{3,3}$ <BRK> \\ 
4 & {\color{black!45} <CONT> $  y_{3,1} \: y_{3,2} \: y_{3,3} $} <SEP> $y_{4,1}  y_{4,2}$ <END> \\
\end{tabular}

 \caption{Illustrated example of sample construction with
 history. Starting from a corpus of ordered sentence pairs (top),
 streaming samples are constructed (bottom) using $h=5$. Past history
 is shown in light gray. Sentence boundary and document
 tokens \cite{junczys2019microsoft} are not counted for the history
 size limit. Notice how, for the last sample, the pair $(\bm x_2, \bm
 y_2)$ is not included in the sample, as the history size limit would
 have otherwise been exceeded on the source
 side.  \label{fig:corpus_ex}}

\end{figure*}

\section{Segmenter System configuration}
The Direct Segmentation system has been trained with the official
implementation
(\url{https://github.com/jairsan/Speech_Translation_Segmenter}). The
following command was used to train the segmenter system:

\small
\begin{verbatim}
python3 train_text_model.py \
 --train_corpus train.$len_$window.txt \
 --dev_corpus  dev.$len_$window.txt \
 --output_folder $out_f \
 --vocabulary $corpus_f/train.vocab.txt \
 --checkpoint_interval 1 \
 --epochs 15 \
 --rnn_layer_size 256 \
 --embedding_size 256 \
 --n_classes 2 \
 --batch_size 256 \
 --min_split_samples_batch_ratio 0.3 \
 --optimizer adam \
 --lr 0.0001 \
 --lr_schedule reduce_on_plateau \
 --lr_reduce_patience 5 \
 --dropout 0.3 \
 --model_architecture ff-text \
 --feedforward_layers 2 \
 --feedforward_size 128 \
 --sample_max_len $len \
 --sample_window_size $window
\end{verbatim}
\normalsize

with the following configurations: \begin{verbatim}(len=11; window=0) 
(len=12; window=1)
(len=13; window=2)
(len=14, window=3)
(len=15, window=4) \end{verbatim}

\end{document}